\documentclass{article}
\usepackage{spconf}
\usepackage{hyperref}       
\usepackage{url}            
\usepackage{booktabs}       
\usepackage{amsfonts}       
\usepackage{nicefrac}       
\usepackage{microtype}      
\usepackage{epsfig}
\usepackage{graphicx}
\usepackage{amsmath}
\usepackage{amssymb}
\usepackage{bm}
\usepackage{amstext}
\usepackage{multirow}
\usepackage{pifont}
\usepackage{array}
\usepackage{indentfirst}
\usepackage[american]{babel}
\usepackage{xcolor,colortbl}
\usepackage[ruled, linesnumbered]{algorithm2e}

\newcommand{\etc}{\textit{etc}}

\newcommand{\ie}{\textit{i}.\textit{e}.}

\title{DMT: Comprehensive Distillation with Multiple Self-supervised Teachers}
\name{
  Yuang Liu\textsuperscript{\rm 1,2}\sthanks{This work was supported by DAMO Academy through DAMO Academy Research Intern Program.} \qquad Jing Wang\textsuperscript{\rm 2,3} \qquad Qiang Zhou\textsuperscript{\rm 2,3} \qquad Fan Wang\textsuperscript{\rm 2,3} \qquad Jun Wang\textsuperscript{\rm 1} \qquad Wei Zhang\textsuperscript{\rm 1}\sthanks{Corresponding author.}
}
\address{
  \textsuperscript{\rm 1}East China Normal University \qquad \textsuperscript{\rm 2}DAMO Academy, Alibaba Group \qquad \textsuperscript{\rm 3} Hupan Lab
}
\begin{document}
%
\maketitle
\begin{abstract}
Numerous self-supervised learning paradigms, such as contrastive learning and masked image modeling, have been proposed to acquire powerful and general representations from unlabeled data. However, these models are commonly pretrained within their specific framework alone, failing to consider the complementary nature of visual representations. To tackle this issue, we introduce Comprehensive \underline{D}istillation with \underline{M}ultiple Self-supervised \underline{T}eachers (DMT) for pretrained model compression, which leverages the strengths of multiple off-the-shelf self-supervised models. 
Our experimental results on prominent benchmark datasets exhibit that the proposed method significantly surpasses state-of-the-art competitors while retaining favorable efficiency metrics. On classification tasks, our DMT framework utilizing three different self-supervised ViT-Base teachers enhances the performance of both small/tiny models and the base model itself. For dense tasks, DMT elevates the AP/mIoU of standard SSL models on MS-COCO and ADE20K datasets by 4.0\%.
\end{abstract}
\begin{keywords}
Distillation, Self-supervised Learning, Multiply Teachers
\end{keywords}

\section{Introduction}

\begin{figure}[!t]
    \centering
    \includegraphics[width=.88\linewidth]{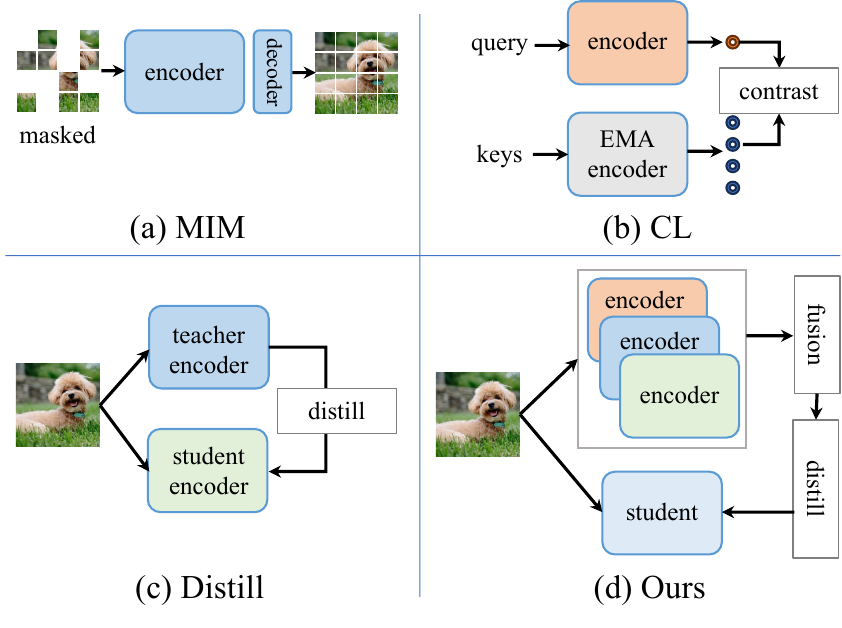}
    \vspace{-3mm}
    \caption{Various paradigms of self-supervised learning, including (a) Masked Image Modeling (MIM), (b) Contrastive Learning (CL), (c) Distillation, and (d) Our DMT.}
    \label{fig:overview}
\end{figure}

Self-supervised learning (SSL)~\cite{zbontar2021barlow,liu2021self,jing2020self} has gained immense popularity in deep learning due to its robust transferability across various vision tasks, including image classification, object detection, and semantic segmentation. Unlike supervised learning, SSL aims to extract rich and task-agnostic representations from massive unlabeled data without requiring costly human annotations. Typically, it involves training an encoder or backbone on a pretraining stage and subsequently fine-tuning the model on downstream tasks. Recently, a profusion of SSL methods have come to light in two dominant branches as shown in Fig.~\ref{fig:overview}: contrastive learning (CL)~\cite{he2020momentum,chen2020improved,zhang2023adclr} and masked image modeling (MIM)~\cite{chen2020simple,he2022masked,chen2022context}. Contrastive learning aims to learn representations by maximizing the agreement between differently augmented views of the same image in latent space, while masked image modeling is to recover the corrupted pixels or features of an image by leveraging the context information. 

However, both contrastive learning and masked image modeling suffer from two critical drawbacks. Firstly, their training processes are time-consuming and resource-intensive, which restricts their usage in individual scenarios. For instance, in research experiments or real-world downstream tasks, we typically adopt only the officially released pretrained model, making it challenging to retrain a customized smaller or new architecture network. Secondly, the pretrained models are often large and cumbersome, unsuitable for deployment on resource-constrained devices. Additionally, while smaller models are friendly to deployment, they do not benefit from the advantages of large-scale pretraining and may experience performance degradation~\cite{ren2022tinymim}. 

\begin{figure*}[!t]
  \centering
  \includegraphics[width=.75\linewidth]{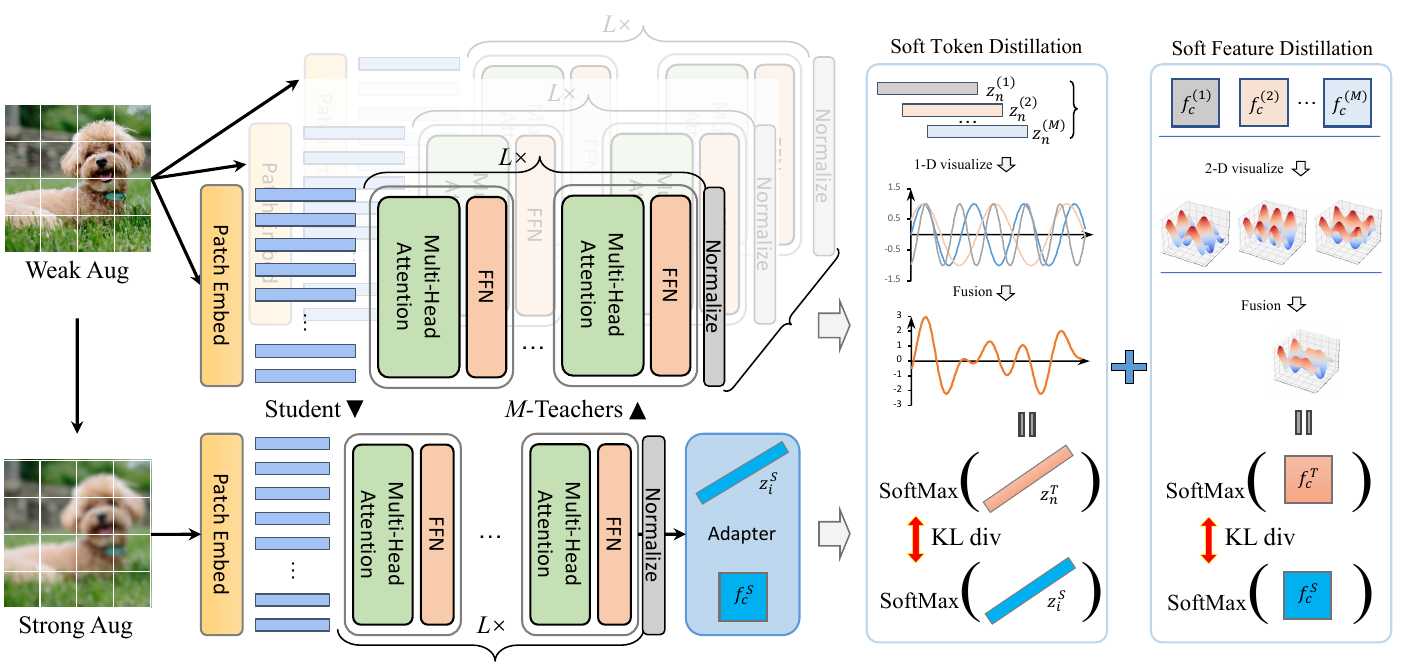}
  \caption{Framework of the proposed DMT. A series of simple and abstract sine wave and surface graph are plotted to represent tokens/features' distribution.}
  \label{fig:framework}
\end{figure*}

Knowledge distillation (KD)~\cite{hinton2015distilling} is a promising technique to address the aforementioned issues. It aims to transfer the knowledge from a large teacher model to a small student model, which can boost the performance and achieve model compression~\cite{bucilua2006model}. Recently, KD has been widely applied to various deep learning tasks, such as image classification~\cite{hinton2015distilling,kim2021self}, detection~\cite{luo2018graph,tang2021humble}, segmentation~\cite{he2019knowledge,liu2023self-decoupling}, large-scale language models (LLM)~\cite{liu2020fastbert}, \etc. However, most existing KD methods are designed for supervised learning on task-specific tasks and cannot be directly applied to SSL. While some researchers have tackled the SSL distillation problem~\cite{touvron2021training,ren2022tinymim}, they only focus on the distillation with a single teacher model. In classification tasks, a group of advanced researches~\cite{you2017learning,liu2020adaptive} have introduced multiple well-trained teachers to guide the training of student network, but they struggle to balance the weights of different teachers in task-specific distillation loss. 

Therefore, in this paper, we propose comprehensive \underline{D}istillation with \underline{M}ultiple self-supervised \underline{T}eachers (DMT) for pretrained model compression, which leverages the strengths of multiple off-the-shelf self-supervised models to individually train customized smaller networks. Our main contributions are summarized as follows:

\textbf{1) Pioneering work in distilling multiple self-supervised teachers.} 
With the emergence of various self-supervised models, there is a wealth of latent information that can be extracted and reused for compact model pretraining instead of learning from scratch. Moreover, due to the different training setups and paradigms, different SSL models tend to learn distinct representations, which can complement each other and build a comprehensive knowledge base. To the best of our knowledge, this represents the first effort to distill from multiple self-supervised teachers. 

\textbf{2) Dual distillation strategy with soft fusion.}
A simple yet efficiency dual distillation approach is proposed to combine the different knowledge of multiply self-supervised teacher and guide the compact student. We remove the isomeric decoders of various teachers and fuse their latent information from two dimensions as soft knowledge, \ie, token fusion distillation and spatial fusion distillation. 

\textbf{3) Significant improvement on both classification and dense tasks.}
We conduct extensive experiments on standard benchmarks, including classification, detection, and segmentation. For ViT-Tiny, DMT outperforms MAE by 3.4\% top-1 accuracy on ImageNet-1K. For ViT-Small with Cascade R-CNN, DMT outperforms iBOT by almost 4\% $\text{AP}^{bb}$ and 3\% $\text{AP}^{mk}$ on MS-COCO dataset.

\section{Methodology}

\subsection{Preliminaries}

\textbf{Vision Transformers.}
Given an RGB image ${\rm x}\in\mathbb{R}^{3\times H\times W}$ as input (where $H, W$ refer to height and width), the top embedding layer of plain ViT~\cite{dosovitskiy2020image} divides it into $N$ non-overlapping patches and then projects each one to a $D$-dimensional token embedding $\{\mathbf{t}_{n} \in \mathbb{R}^{D}\}^{N}_{n=1}$. When setting the patch size as $p$, $N = (H/p)\times (W/p)$. A class token $\mathbf{t}_0$ is subsequently prepended to the sequence of token embeddings to capture global information. The sequence of tokens is then fed into the $L$-layer ViT encoder $\mathcal{E}$ to obtain the final representation 
\begin{equation}
  \mathbf{Z} = \mathcal{E}\left(\left[\mathbf{t}_0, \mathbf{t}_1, \cdots, \mathbf{t}_N \right]\right) \,, 
  \mathbf{Z} \in \mathbb{R}^{(N+1)\times D} \,.
\end{equation}
$\mathbf{Z} = \{\mathbf{z}_n\}_{n=0}^{N}$ is a sequence consisting of $(N+1)$ token embeddings. 

\textbf{Self-supervised learning with ViTs.}
Whether using CL or MIM self-supervised frameworks, they generally introduce an auxiliary decoder $\mathcal{D}$ following the backbone/encoder network, which can be formulated as $\mathbf{y} = \mathcal{D} \circ \mathcal{E}\left(\left[\mathbf{t}_0, \mathbf{t}_1, \cdots, \mathbf{t}_N \right]\right) $.  And the contrastive or reconstruction loss is built with $\mathbf{y}$ (pixel, features, labels, \etc) output by the customized decoder. In this work, we obtain pretrained teacher models using this encoder-decoder learning framework, but remove the $\mathcal{D}$ and only utilize the encoder $\mathcal{E}$ for distillation-based pretraining, which can reduce the training cost and unify the latent space $\mathbf{z}$ of encoders pretrained by different SSL frameworks. 

\subsection{DMT with Dual Soft Fusion Distillation}

Fig.~\ref{fig:framework} depicts the framework of the proposed DMT. We fuse the various knowledge from multiple teachers and perform distillation from two aspects: token-wise and spatial feature-wise fusion distillation. 

Assuming there are $M$ pretrained teachers $\{\mathcal{T}^{(m)}\}^{M}_{m=1}$ with weak data augmentation, we can denote the latent token sequence of the $m$-th teacher as $\mathbf{Z}^{(m)} \in \mathbb{R}^{(N+1)\times D}$, where $\mathbf{z}^{(m)}_n$ is the $n$-th token embedding of the $m$-th teacher. All teachers share the same embedding dimension $D$. In order to obtain the augmented token representations, each of which is $\mathbf{z}^S \in \mathbb{R}^{D'}$, we introduce a strong image augmentation strategy to the student network $\mathcal{S}$. In particular, random resized cropping and random horizontal flipping are used for both teachers and student, while color jitter is only applied to the student. To match the embedding dimensions of the student and teachers, we employ a lightweight auxiliary adapter to project the $\mathbf{z}^S : \mathbb{R}^{D'} \rightarrow \mathbb{R}^{D}$. The pretrained Layer Normalization (LN)~\cite{ba2016layer} at the top of each teacher and student are preserved to normalize the representation space. 

\textbf{Soft Token Fusion Distillation.}
With normalized embeddings $\mathbf{Z} \in [-1, 1]$, the tokens can be regarded as undulating wave distribution. With the enhancement or reduction of waves at different positions, the signals are mixed adaptively. The agreement of multiple teachers would be strengthened and the disagreement would be weakened. Therefore, aggregating the teachers' token embeddings $\{\mathbf{z}^{(m)}_n\}_{m=1}^{M}$ at the same position $n$ to an enhanced one is a straightforward way for transferring rich information to the student, $\mathbf{z}^T_n = \sum_{m=1}^{M} \mathbf{z}^{(m)}_n $.
Here, $\mathbf{z}^T_{n}$ represents the aggregated token embedding of the $n$-th token from multiple teachers. We formulate the token fusion distillation (TFD) loss as the Kullback-Leibler (KL) divergence between the student and aggregated teacher token embeddings after a softmax function $\varphi$: 
\begin{equation}
  \mathcal{L}_{TFD} = \frac{1}{N+1}\sum_{n=0}^{N} \sum_{j=0}^{D-1} \varphi(\mathbf{z}^S_{n})_j \log \frac{\varphi(\mathbf{z}^S_{n})_j}{\varphi(\mathbf{z}^T_{n})_j} \,,
\end{equation}
where $\varphi(\mathbf{z}_{n})_j = \frac{e^{\mathbf{z}_{n,j}}}{\sum_{d=1}^{D} e^{\mathbf{z}_{n,d}}}$. It's important to note that the KL divergence can capture the distributions and reflect the latent relation between the items in each token embedding, which is more effective than the point-to-point Euclidean distance. We conduct an ablation study about this in our experiments. 

\textbf{Soft Feature Fusion Distillation.}
Besides token-wise fusion and distillation, we can gather the sequence of all tokens (excluding the class token) denoted as $\mathbf{Z}_{[1:N]} = \{\mathbf{z}_n | \mathbf{z}_n\in \mathbb{R}^{D}\}_{n=1}^{N}$, and reconstruct it into a 3-D feature representation $\mathbf{F} \in \mathbb{R}^{D \times H' \times W'}$, which can transfer channel-wise spatial information. Note that $H' = H/p$ and $W' = W/p$ indicate the height and width of each feature map. The $c$-th channel of student feature map is denoted as $\boldsymbol{f}^S_c$, and $\boldsymbol{f}^{(m)}_c$ is the $c$-th channel of the $m$-th teacher feature map. With these insights, we can extend the 1-D token fusion distillation to 2-D feature fusion distillation (SFD) to capture the spatial knowledge from teachers' feature maps. 
For simplicity, we reshape the student/teacher feature map $\boldsymbol{f}_c: \mathbb{R}^{H' \times W'} \rightarrow \mathbb{R}^{N}$, and obtain the aggregated knowledge from teachers by $\boldsymbol{f}^{T}_c = \sum_{m=1}^{M} \boldsymbol{f}^{(m)}_c$. 
The spatial fusion distillation (SFD) can be expressed as channel-wise KL difference between the student and aggregated teacher feature maps:
\begin{equation}
  \mathcal{L}_{SFD} = \frac{1}{D}\sum_{c=0}^{D-1} \sum_{n=0}^{N-1} \varphi(\boldsymbol{f}^S_c)_n \log \frac{\varphi(\boldsymbol{f}^S_c)_n}{\varphi(\boldsymbol{f}^T_c)_n} \,.
\end{equation}
The total objective function of our DMT framework is 
\begin{equation}
  \mathcal{L} = \mathcal{L}_{TFD} + \mathcal{L}_{SFD} \,.
\end{equation}
Here we do not introduce any hyper-parameters as previous KD methods.

\begin{table}[t]
  \centering
  \caption{Finetune accuracy (\%) on ImageNet-1K val set. }
  \resizebox*{\columnwidth}{!}{
    \begin{tabular}{lc|c|c}
    \toprule
    Method & Teacher & Epochs & Top-1 \\
    \midrule
    \textit{\textbf{ViT-T  (5M)}} &       &       &  \\
    DeiT  & Label & 300   & 72.2 \\
    MAE   & Pixel & 1600  & 71.6 \\
    MoCoV3 & EMA   & 1600  & 73.3 \\
    TinyMIM & MAE ViT-B & 300   & 74.6 \\
    TinyMIM & iBOT ViT-B & 300   & 74.6 \\
    TinyMIM & MoCoV3 ViT-B & 300   & 74.5 \\
    DMT (Ours) & MAE/MoCoV3/iBOT & 300   & \textbf{75.0} \\
\hline
\textit{\textbf{ViT-S (22M)}} &       &       &  \\
    DeiT  & Label & 300   & 79.9 \\
    MAE   & Pixel & 1600  & 80.6 \\
    MoCoV3 & EMA   & 1600  & 81.4 \\
    DINO  & EMA   & 1600  & 81.5 \\
    BeiT  & DALL-E & 3200  & 81.6 \\
    TinyMIM & MAE ViT-B & 300   & 81.7 \\
    TinyMIM & MoCoV3 ViT-B & 300   & 81.5 \\
    TinyMIM & iBOT ViT-B & 300   & 81.5 \\
    DMT (Ours) & MAE/MoCoV3/iBOT & 300   & \textbf{82.2} \\
\hline    
\textit{\textbf{ViT-B (86M)}} &       &       &  \\
    MAE   & Pixel & 300   & 82.8 \\
    MoCoV3 & EMA   & 800   & 83.2 \\
    iBOT  & Pixel/EMA & 300   & 83.2 \\
    DMT (Ours) & MAE/MoCoV3/iBOT & 300   & \textbf{83.7} \\
    \bottomrule
    \end{tabular}%
  }
  \label{tab:cls}%
\end{table}%

\section{Experiments}

\subsection{Experimental setup} 
We employ ViT-B models pretrained 300 epochs by three SSL methods as teachers, including MAE~\cite{he2022masked}, MoCoV3~\cite{chen2021empirical}, and iBOT~\cite{zhou2022ibot}. All student models are pretrained 300 epochs on the ImageNet-1K training set. The batch size is set to 4096, and the batch learning rate is initialized as $1.5\times 10^{-4}$, which is decayed by a cosine schedule with a warm-up for 15 epochs. We utilize AdamW~\cite{loshchilov2017decoupled} optimizer with a weight decay of 0.05. Random resized cropping and random horizontal flipping were used for both teachers and students, while color jitter was only applied to the students. 

\subsection{Results}

\textbf{Main results.} 
Tab.~\ref{tab:cls} reports top-1 accuracy on the validation set. Apart from official TinyMIM for MAE, we extend it to MoCoV3~\cite{chen2021empirical} and iBOT~\cite{zhou2022ibot} as two strong baselines of single-teacher KD. For ViT-S, DMT surpasses both standard SSL methods and extended TinyMIM by 0.5\% top-1 accuracy, where the student achieves 98.8\% performance of the cumbersome teachers with only about 25\% parameters. In addition, we conduct ``base to base'' distillation for ViT-B, where the teachers and student are with the same architecture. DMT achieves 83.7\% top-1 accuracy, improving MAE and iBOT by 0.9\% and 0.5\%, respectively. 

\textbf{Transfer Learning Tests.} 
We fine-tune the pretrained ViT-S backbone with Mask R-CNN~\cite{he2017mask} on MS-COCO and UPerNet\cite{lin2017feature} on ADE20K, and present the results in Tab.~\ref{tab:dense}. We calculate box average precision ($\text{AP}^{bb}$) and mask average precision ($\text{AP}^{mk}$) with a confidence of 0.5:0.95 for object detection and instance segmentation on COCO, respectively. DMT outperforms three extended TinyMIM methods up to 2.1\% mIoU and 1.6\% mAcc on ADE20K. There is a margin of 1.7\% mIoU, 1.2\% $\text{AP}^{bb}$ and 1.0\% $\text{AP}^{mk}$ between DMT and the newest SSL method PQCL~\cite{zhang2023pqcl}. 

\begin{table}[t]
    \centering
    \caption{Results on ADE20K and MS-COCO with ViT-S. TinyMIM is abbreviated as ``TM''. }
    \resizebox*{\columnwidth}{!}{
      \begin{tabular}{l|cc|cc}
      \toprule
      \multirow{2}{*}{Method} & \multicolumn{2}{c|}{ADE20K} & \multicolumn{2}{c}{MS-COCO} \\
  \cline{2-5}          & mIoU  & aAcc  & $\text{AP}^{bb}$ & $\text{AP}^{mk}$ \\
      \midrule
      DeiT~\cite{touvron2021training}  & 43.1  & -     & 43.1  & 38.4 \\
      MAE~\cite{he2022masked} & 42.8 & - & - & - \\
      MoCoV3~\cite{chen2021empirical} & 43.9  & -     & 39.8  & 37.1 \\
      iBOT~\cite{zhou2022ibot}  & 44.1  & 81.4  & 42.6  & 39.0 \\
      DINO~\cite{caron2021emerging}  & 42.3  & 80.4  & 40.8  & 37.3 \\
      ADCLR~\cite{zhang2023adclr} & 44.2  & 81.8  & 43.8  & 39.2 \\
      PQCL~\cite{zhang2023pqcl}  & 45.2  & 81.9  & 43.1  & 39.3 \\
      TM-MAE~\cite{ren2022tinymim} & 44.8  & 81.9  & 41.4  & 38.1 \\
      TM-MoCoV3~\cite{ren2022tinymim} & 45.3  & 82.0  & 42.1  & 38.9 \\
      TM-iBOT~\cite{ren2022tinymim} & 46.0  & 82.3  & 43.1  & 39.6 \\
      \textbf{DMT (Ours)} & \textbf{46.9} & \textbf{82.9} & \textbf{44.3} & \textbf{40.3} \\
      \bottomrule
      \end{tabular}%
    }
    \label{tab:dense}%
\end{table}%

\subsection{Ablation Study}

\textbf{Effect of different distillation losses.}
We conduct ablation experiments on ADE20K with ViT-T to explore the effect of different distillation loss functions, as shown in Table~\ref{tab:ab_loss}. Firstly, we extend TinyMIM~\cite{ren2022tinymim} to a multi-teacher version (named m-TinyMIM) by simply averaging the three groups of relational distillation losses. We find that the mean loss function is difficult to be optimized due to the inconsistent guideline from three teachers, and the results are obviously bad. We then replace the KL divergence loss with the MSE loss, which achieves lower performance. Finally, we verify the effectiveness of the proposed two distillation losses, \ie, $\mathcal{L}_{TFD}$  and $\mathcal{L}_{SFD}$. Both of them can outperform TinyMIM-MAE over 2\% mIoU, and the combination of them can further improve the performance by 2.8\% mIoU.

\begin{table}[t]
  \centering
  \caption{Performance of ViT-T on ADE20K with different distillation losses. }
  \resizebox*{\columnwidth}{!}{
    \begin{tabular}{ccccc|ccc}
    \toprule
    $\mathcal{L}_{TFD}$   & $\mathcal{L}_{SFD}$   & KL    & MSE   & m-TinyMIM & mIoU  & aAcc  & mAcc \\
    \midrule
    \multicolumn{5}{c|}{TinyMIM (baseline)}         & 39.2  & 79.9  & 50.2 \\
    \hline
          &       &       &       & \checkmark   & 34.5  & 78.2  & 45.3 \\
    \checkmark   & \checkmark   &       & \checkmark   &       & 41.7  & 80.7  & 53.0 \\
    \checkmark   &       & \checkmark   &       &       & 41.7  & 80.8  & 53.2 \\
          & \checkmark   & \checkmark   &       &       & 41.6  & 80.7  & 52.8 \\
    \checkmark   & \checkmark   & \checkmark   &       &       & \textbf{42.0} & \textbf{81.2} & \textbf{53.7} \\
    \bottomrule
    \end{tabular}
  }
  \label{tab:ab_loss}
\end{table}

\textbf{Combination of teachers.}
To evaluate the contribution of multiple teachers, we try different combinations of them in pretraining distillation. Table~\ref{tab:ab_teacher} depicts the fine-tuning results on ADE20K dataset. We report the metrics improvement in the parenthesis. For two-teacher DMT, we can observe that the combination of MAE~\cite{wei2022masked} and MoCoV3~\cite{chen2021empirical} outperforms both TinyMIM-MAE and TinyMIM-MoCoV3, which is 0.5\% mIoU higher than the single teacher. The combination of all three teachers bring the best performance, surpassing TinyMIM-MAE by 1.7\% mIoU. More teachers can provide more diverse knowledge, which is beneficial to the performance improvement. But more teachers also bring unbearable memory, power cost but limited improvement. Therefore, choosing three teachers is reasonable in our experiments.

\begin{table}[htbp]
  \centering
  \caption{Performance of ViT-T on ADE20K with different teacher combinations.}
  \resizebox*{\columnwidth}{!}{
    \begin{tabular}{ccc|lll}
    \toprule
    MAE & MoCoV3 & iBOT & mIoU  & aAcc  & mAcc \\
    \midrule
    \checkmark   &       &       & 39.2  & 79.9  & 50.2 \\
          & \checkmark   &       & 40.3  & 80.5  & 51.5 \\
          &       & \checkmark   & 40.0  & 80.4  & 51.0 \\
    \checkmark   & \checkmark   &       & 40.8 (+0.5)  & 80.4 (-0.1)  & 52.0 (+0.5) \\
    \checkmark   &       & \checkmark   & 40.2 (+0.2)  & 80.1 (-0.3)  & 51.4 (+0.4) \\
    \checkmark   & \checkmark   & \checkmark   & \textbf{42.0 (+1.7)} & \textbf{81.2 (+0.7)} & \textbf{53.7 (+2.2)} \\
    \bottomrule
    \end{tabular}
  }
  \label{tab:ab_teacher}
\end{table}

\section{Conclusion}

This paper presents DMT, a novel approach that utilizes multiple off-the-shelf self-supervised models to boost the performance of smaller ViT models on downstream tasks. To aggregate comprehensive knowledge from different self-supervised teachers, we propose two distillation strategies, namely token fusion distillation (TFD) and spatial fusion distillation (SFD). Extensive experiments on three common vision tasks demonstrate that our method significantly improves the performance of smaller ViT models with distilled comprehensive information. 

\vfill\pagebreak

{\small
\bibliographystyle{IEEEbib.bst}
\bibliography{refs.bib}
}

\end{document}